\documentclass[11pt]{article}

\usepackage[margin=1in]{geometry} 
\usepackage{amsmath,amssymb,amsfonts}
\usepackage{graphicx}
\usepackage{textcomp}
\usepackage{xcolor}
\usepackage{multirow}
\usepackage{bm}
\usepackage{color}
\usepackage{lipsum}
\usepackage{algorithmic}
\usepackage{tabularx}
\usepackage{booktabs}
\usepackage{authblk}

\usepackage[backend=bibtex,style=numeric,sorting=none]{biblatex}
\title{Bayes-DIC Net: Estimating Digital Image Correlation Uncertainty with Bayesian Neural Networks}

\author[1]{Biao Chen}
\author[3]{Zhenhua Lei}
\author[2]{Yahui Zhang\thanks{Corresponding author, email: yahui.zhang@northwestern.edu}}
\author[4]{Tongzhi Niu}

\affil[1]{Department of Mechanical Engineering, University of Michigan, Ann Arbor, MI 48109, USA}
\affil[2]{Department of Mechanical Engineering, Northwestern University, Evanston, IL 60208, USA}
\affil[3]{School of Mechanical Science and Engineering, Huazhong University of Science and Technology, Wuhan 430074, Hubei, China}
\affil[4]{School of Artificial Intelligence and Robotics, Hunan University, Changsha 410012, Hunan, China}

\date{} 

\begin{document}

\maketitle

\begin{abstract}
This paper introduces a novel method for generating high-quality Digital Image Correlation (DIC) dataset based on non-uniform B-spline surfaces. By randomly generating control point coordinates, we construct displacement fields that encompass a variety of realistic displacement scenarios, which are subsequently used to generate speckle pattern datasets. This approach enables the generation of a large-scale dataset that capture real-world displacement field situations, thereby enhancing the training and generalization capabilities of deep learning-based DIC algorithms.
Additionally, we propose a novel network architecture, termed Bayes-DIC Net, which extracts information at multiple levels during the down-sampling phase and facilitates the aggregation of information across various levels through a single skip connection during the up-sampling phase. 
Bayes-DIC Net incorporates a series of lightweight convolutional blocks designed to expand the receptive field and capture rich contextual information while minimizing computational costs.
Furthermore, by integrating appropriate dropout modules into Bayes-DIC Net and activating them during the network inference stage, Bayes-DIC Net is transformed into a Bayesian neural network. This transformation allows the network to provide not only predictive results but also confidence levels in these predictions when processing real unlabeled datasets. 
This feature significantly enhances the practicality and reliability of our network in real-world displacement field prediction tasks. Through these innovations, this paper offers new perspectives and methods for dataset generation and algorithm performance enhancement in the field of DIC.
\end{abstract}

\textbf{Keywords:} Digital Image Correlation (DIC), Deep Learning, Network Architecture, Bayesian Neural Networks.

\section{Introduction}




\section{Introduction}

The Digital Image Correlation (DIC) method is a sophisticated optical technique for non-contact measurement of global displacements and strain fields utilizing visible light \cite{hild2006digital} \cite{reu2012introduction}. This technique is capable of quantifying the displacement field of a specimen by employing a camera as the sole sensing device \cite{gorszczyk2019application}. Compared to alternative measurement techniques, DIC offers several distinctive advantages, including the cost-effectiveness of the measurement apparatus and the capability to conduct comprehensive global assessments \cite{blikharskyy2022review}. These benefits have established DIC as a highly invaluable measurement tool in the field of experimental mechanics.

As for the operational mechanics of traditional DIC algorithms, they function by partitioning the target image into discrete subregions and correlating these with a predefined reference image \cite{sutton1983determination, peters1982digital, pan2009two}. This process relies on the establishment of robust correlation criteria to compute the displacement field accurately \cite{pan2011recent}.
Since the seminal introduction of the DIC method in the 1980s, it has been the subject of extensive scholarly research. Research efforts have been dedicated towards enhancing its performance, precision, and stability, as well as broadening the spectrum of its applications \cite{blikharskyy2022review}.

In recent years, the rapid advancement of deep learning technology in the field of computer vision has demonstrated significant potential in digital image correlation (DIC) \cite{yang2022deep}. The powerful feature extraction and pattern recognition capabilities of deep learning methods enable the automatic learning and extraction of complex patterns and features from images. This is particularly beneficial for tasks such as DIC, which require precise measurement of object deformation. Currently, state-of-the-art DIC algorithms have been applied in various fields, including thermal deformation measurement, object pose measurement, and underwater deformation measurement \cite{sutton2009image}. However, as deep learning-based DIC methods receive increasing attention as a research hotspot, several challenges are becoming more evident \cite{boukhtache2021deep}.

Firstly, the quality of datasets used for training deep learning models in DIC is often inadequate. The collection and annotation of real-world datasets are typically costly and time-intensive processes \cite{pan2011recent}. Moreover, obtaining high-quality data is particularly challenging under specific conditions, such as extreme environments or difficult-to-access locations. Due to these limitations, much research has shifted towards the use of manually generated speckle image datasets for training neural networks \cite{yang2022deep}. Although these datasets can simulate real image features to some extent, they often lack the complexity and diversity of real-world scenarios, which may result in limited generalization capabilities of the trained models. Therefore, how to efficiently generate high-quality and realistic data integration is the first problem to be solved for deep learning-based DIC methods.

Secondly, during the material testing process, the material undergoes continuous deformation, which requires the DIC method to comprehensively and accurately capture the deformation process \cite{hild2006digital}. This requirement places extremely high demands on the feature extraction capability of DIC networks. Therefore, proposing neural networks with higher feature extraction and generalization capabilities has become a key research focus.

Finally, in real situations, the accuracy of network predictions is difficult to estimate \cite{gal2016dropout}. Currently, researchers emphasize training networks with high predictive performance on artificially generated datasets. These datasets are manually labeled, facilitating straightforward accuracy assessment. However, real-world data collected during measurements differs from the training data and lacks definitive labels. Consequently, when researchers infer their trained networks on real unlabeled datasets, they struggle to ascertain the accuracy and reliability of neural network predictions. This difficulty undermines confidence in the effectiveness of their methods for real-world measurement tasks.

To solve the above challenges, this paper first proposes a dataset generation method based on non-uniform B-spline surfaces. By randomly controlling the coordinates of points, a displacement field encompassing diverse real-world scenarios is generated. This field serves as a label for generating speckle image pairs, thereby enhancing the training of neural network algorithms. 
Secondly, a new neural network structure is proposed, retaining the Encoder-Decoder structure while incorporating  the concept of Bilateral architectures to extract both detailed low-level and high-level semantic information. Unlike traditional approaches that add extra branches, this architecture integrates different feature extraction layers within the same path: shallow layers capture detailed information, while deeper layers focus on global content. Besides, we fuse detailed information and global information through jump connection structures. This design not only preserves the multi-scale feature extraction and fusion characteristics but also minimizes memory usage typically incurred by multiple skip connections, enabling higher inference speeds.
Finally, by integrating dropout structures into network and activating them during inference, the neural network is transformed into a Bayesian neural network. Leveraging Bayesian characteristics allows the network to provide confidence estimates alongside predictions during inference on real, unlabeled datasets. The proposed neural network algorithm is thus named Bayes-DIC Net.

In summary, our main contributions are as follows:

1) This paper presents a high-quality DIC dataset generation method based on non-uniform B-spline surfaces. We generate displacement fields containing various possible real displacement situations by randomly generating control point coordinates. These fields are then utilized to generate speckle pattern datasets. 
This approach allows us to produce a comprehensive dataset that captures diverse real-world displacement field scenarios. Consequently, this dataset enhances the training of deep learning-based DIC algorithms and improves their ability to generalize across different scenarios.

2) We propose a novel network architecture, dubbed Bayes-DIC Net, which extracts information at different levels in stages during the down-sampling stage and facilitates the aggregation of information at various levels through one skip connection during the up-sampling stage. Besides, A series efficient convolution blocks are designed for Bayes-DIC Net, which can enhance the receptive field and capture rich contextual information.

3) By adding some appropriate dropout modules into Bayes-DIC Net and activating them during the network inference stage, Bayes-DIC Net becomes a Bayesian neural network. According to the special properties of Bayesian neural networks, they can provide both prediction results and associated confidence when reasoning on real unlabeled datasets. This will greatly expand the practicality and reliability of our network in real displacement field prediction tasks.


\section{Related Works}

\subsection{Traditional Digital Image Correlation}
Digital Image Correlation (DIC) is an advanced optical technique that enables the extraction of comprehensive field data related to shape, motion, and deformation from images \cite{hild2006digital}. This non-contact method enhances the robustness and accuracy of measurements by obviating the need for physical interaction with the specimen under investigation. Since its inception in the early 1980s, DIC algorithms have undergone substantial refinement, yielding improvements in both computational efficiency and measurement precision \cite{sutton1983determination}.

Early foundational work established two primary correlation criteria within DIC: the Sum of Squared Differences (SSD) and the cross-correlation (CC) functions \cite{sutton1983determination} \cite{sutton1986application}. Building upon these foundations, a variety of sophisticated correlation metrics have been developed, including the Zero-Normalized Cross-Correlation (ZNCC) and the Parametric Sum of Squared Difference (PSSD) \cite{pan2011recent}. These criteria are pivotal in the calculation of the displacement field, where the objective is to ascertain the image similarity through the optimization of the CC coefficient or the minimization of the SSD coefficient.

In the pursuit of sub-pixel displacement resolution, a multitude of registration algorithms have been proposed, generally categorized into two classes: local subset-based methods and global (continuum) approaches. Local subset-based methods leverage interpolation of grayscale pixel values \cite{peters1982digital} or utilize a correlation matrix within each subset \cite{chen1993digital}. They may also employ iterative calculations to resolve non-linear mapping parametric vectors \cite{bruck1989digital} and spatial gradients \cite{pan2005plane}, or directly identify the peak of a statistical similarity function \cite{ronneberger2015u}. The parallel processing capability of subset-based methods greatly expedites computation; however, they may yield discontinuities at subset boundaries, leading to potential noise in the resulting strain field data.

Conversely, global (continuum) methods conceptualize the displacement field across the entire image domain, employing shape functions and finite element methods to resolve the measurements \cite{sun2005finite}. These methods preserve compatibility and are adept at capturing heterogeneous deformation within localized regions. However, they fall short of the superior precision and efficiency demonstrated by subset-based techniques \cite{wang2016subset}.

\subsection{Digital Image Correlation Based on Deep Learning}
The advent of deep learning has catalyzed significant advancements across a myriad of computer vision applications, ranging from image classification \cite{krizhevsky2017imagenet}, object detection \cite{ouyang2016deepid}, to 3D reconstruction \cite{chen2018ps}. Its applicability has extended to optical flow estimation, a task involving the extraction of displacement fields from pairs of images \cite{zayouna2011local}. Methods founded on Convolutional Neural Networks (CNNs) have demonstrated superiority over traditional optical flow techniques, particularly in aspects of accuracy and computational efficiency \cite{hur2020optical}. By designing networks with multiple convolutional and deconvolutional layers, complemented by appropriate pooling and activation functions \cite{ilg2017flownet}, CNNs exhibit an exceptional capability to delineate optical flow fields with sub-pixel precision across image pairs, even in scenarios characterized by significant displacements \cite{sun2018pwc}.

The robust framework provided by deep learning has engendered interest in its application within the domain of Digital Image Correlation (DIC). Initial endeavors by researchers \cite{min2019strain} attempted to integrate deep learning techniques with traditional DIC processes. Despite these efforts, challenges pertaining to computational accuracy and efficiency were observed. In response, the focus has shifted towards developing end-to-end deep learning models specialized for the measurement of speckle displacement fields. Researchers\cite{boukhtache2021deep} re-purposed existing deep learning optical flow architectures, adjusting them through retraining and fine-tuning with computer-simulated speckle image datasets. This led to the formulation of StrainNet, a model purported to achieve displacement field measurement accuracy comparable to conventional DIC techniques. However, it is notable that the dataset employed for this method primarily encompassed sub-pixel level random displacements, which may not accurately reflect real-world conditions. This limitation implies that the network might only be proficient in measuring displacement fields of less than one pixel, falling short of practical measurement specifications. Subsequently, researchers \cite{yang2022deep} generated synthetic datasets leveraging affine transformations and Gaussian functions to fabricate displacement fields serving as ground truth. These datasets were then used to train a deep learning-based DIC method called DeepDIC, which includes two independent convolutional neural networks for end-to-end displacement and strain field measurements. This development highlights the ongoing evolution of deep learning applications within the DIC field, underscoring a progressive shift towards models that aim to address the practical challenges encountered in traditional DIC methods.

\subsection{Bayesian Neural Network}
Within the domain of deep learning, Bayesian Neural Networks (BNNs) \cite{magris2023bayesian} provide a probabilistic framework for model interpretation by estimating distributions over the models' weights, which closely align with the principles of Gaussian processes. Nonetheless, the extensive number of parameters in these networks presents a significant challenge in modeling a distribution over the kernels, thereby escalating the computational requirements.

Yarin \cite{gal2016dropout} presented a compelling proposition that the application of dropout during the training phase of neural networks can serve as a Bayesian approximation. This perspective facilitates the mitigation of heightened computational complexity and the potential decrement in test accuracy. Then, Yarin introduced a novel architectural framework for dropout in Convolutional Neural Networks \cite{gal2015bayesian}. Accordingly, the training regimen of such networks is redefined as an approximate Bernoulli variational inference within the context of BNNs. The assessment of these models is achieved through the approximation of the predictive posterior distribution, a method termed Monte Carlo dropout, applied during the testing phase.

In this research, we incorporate Dropout into a specifically designed neural network for predicting displacement fields, thereby constructing a Bayesian neural network. As a result, it becomes feasible to ascertain the confidence level of the network in its predictions of the displacement field during the inference phase. This enhancement renders our network more amenable for applications in industrial applications, offering a robust approach for predictive modeling in environments where accuracy and reliability are paramount.

\section{Methodology}


\subsection{DIC Dataset Generation}
Various datasets have played a significant role in advancing the development of deep learning. The availability of vast and diverse datasets has greatly propelled the progress of deep learning across various domains. Hence, to train deep learning-based DIC algorithms of superior performance, access to high-quality datasets suitable for the task is of paramount importance. In the context of material testing, the deformation exhibited by materials prior to catastrophic failures, such as fracture, is often characterized by global, continuous, and irregular patterns. Consequently, a high-quality dataset must adhere to these three fundamental criteria. To this end, we propose a method for generating high-quality DIC datasets based on non-uniform B-spline surfaces. This approach aligns with the essential criteria of global, continuous, and irregular deformation patterns, facilitating more effective training of DIC algorithms.

\begin{figure}[h]
\centering
\includegraphics[width=1\textwidth]{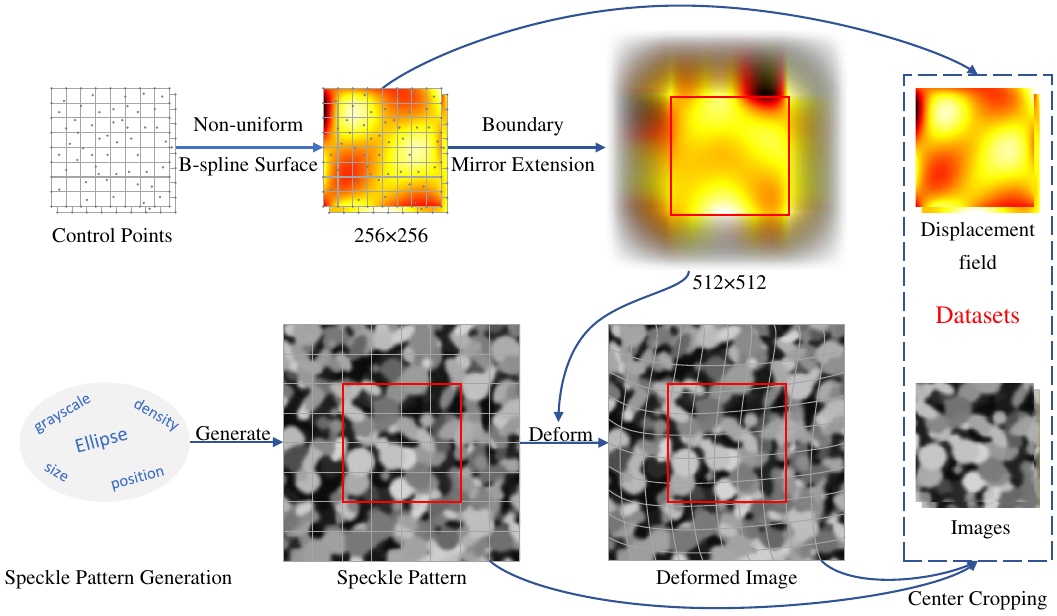}
\caption{Schematic of the dataset generation workflow.}\label{fig1}
\end{figure}

As illustrated in Figure 1, the specific process of the proposed method is as follows: Initially, a speckle pattern is algorithmically generated to serve as the imagery for the dataset. Subsequently, coordinates of control points are randomly generated to create a completely random non-uniform B-spline surface, which is employed as the displacement field for this synthetic dataset. Finally, the generated displacement field is applied to the speckle pattern, resulting in a deformed speckle pattern and thus completing the dataset. With this methodology, the present study has generated a substantial dataset comprising 12,500 image pairs (10,000 pairs for training and 2,500 pairs for testing). This dataset encompasses as wide a variety of realistic displacement fields as possible, thereby enhancing its utility in training deep learning-based DIC algorithms and improving their generalization capabilities. The detailed procedure for dataset generation is outlined below:


Firstly, speckle pattern images are generated by stacking ellipses with random sizes and gray-scale values. We first divide the grayscale values into three segments, namely 0.08-0.38, 0.38-0.68, and 0.68-0.98. Each segment contains 500-4000 ellipses with random sizes within a frame size of 512 × 512. For each sample in the dataset, a unique and random speckle pattern is created, so there is no same speckle images.

In order to generate a displacement field, we create a non-uniform B-spline surface and use the z-coordinate of each point on the surface as the displacement magnitude in the x or y direction. Prior to the generation of the non-uniform B-spline surface, it is necessary to obtain the coordinates of the control points. To this end, we first determine that the number of control points will be a 9x9 grid, and we fix the outermost control points at the boundary of the surface. Specifically, the x or y coordinates of the peripheral control points in the 9x9 control point matrix are fixed at either 0 or 512 to prevent excessively large z-coordinate values at the surface boundary. Furthermore, for the unrestricted x, y, and z coordinates, we randomly generate the x and y coordinates within the surface size range of [0, 256] and [0, 256], respectively, and we constrain the range of the z-coordinates within [-5, 5].

Finally, we utilize an interpolation algorithm to treat the generated non-uniform B-spline surface as the displacement field, thereby obtaining images of the deformed speckle pattern. Given that the speckle images are of size 512x512, while the non-uniform B-spline surface is 256x256 in size, we initially expand the surface to a 512x512 dimension using a mirroring algorithm. This expansion ensures a match with the speckle images. Subsequently, after the interpolation process is completed, the center of the deformed image is cropped to a dimension of 256x256 to produce the final dataset. Such a methodology prevents the issue of missing boundaries in the deformed speckle images, thereby yielding a higher quality dataset.

The high-quality dataset generation method described above possesses the following advantages:

1. The generation of displacement fields via non-uniform B-spline surfaces directly ensures the continuity of the displacement field, aligning with the characteristics of displacement fields observed in material testing processes.

2. By randomly generating the coordinates of control points for non-uniform B-spline surfaces, it is ensured that the displacement fields within the dataset are random. This enhances the network's ability to generalize across various continuous displacement fields.

3. This method of dataset generation is computationally efficient. Coupled with multi-threaded data generation techniques, it allows for the rapid production of large volumes of data. This facilitates a more comprehensive inclusion of possible displacement field scenarios, resulting in a dataset that more closely mirrors real-world conditions.

\begin{figure}[h]
\centering
\includegraphics[width=1\textwidth]{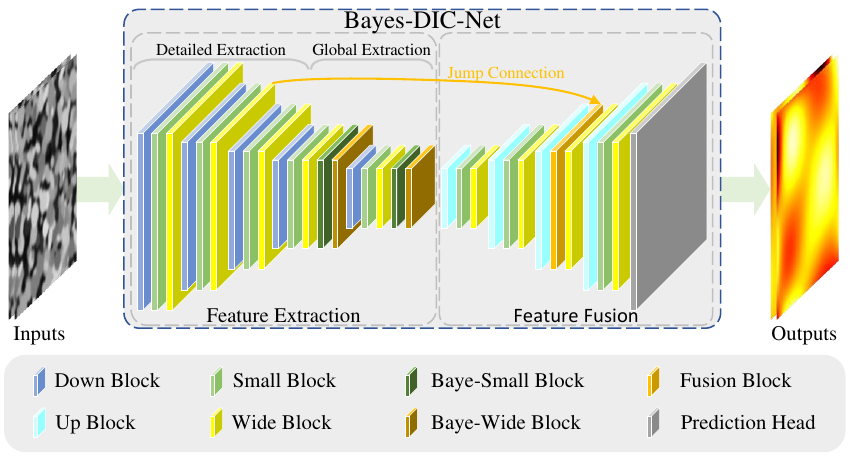}
\caption{The architecture of the proposed network. During the Feature Extraction stage, the feature maps with darker colors correspond to higher levels of information.}\label{fig2}
\end{figure}

\subsection{DIC Network Architecture Design}
\subsubsection{Overview}
As depicted in Figure 2, the proposed Bayes-DIC Net is comprised of two distinct parts, namely feature extraction and feature fusion. We argue that both the low-level detailed displacement and high-level global displacement are important for the displacement prediction. Therefore, the feature extraction stage consists of two stages: detail extraction and global extraction. These stages employ different stacking modes of down-sampling module (Down Block) and feature extraction module (Small Block and Wide Block). The aim of detail extraction is to extract low-level detailed information more effectively, whereas the goal of global extraction is to capture high-level global information more precisely. For the detail extraction stage, we utilize the stacking of Down Block, Small Block, and Wide Block. On the other hand, to rapidly expand the receptive field in the global extraction stage, we use the stacking mode of Down Block, two Small Blocks, and two Wide Blocks. In the feature fusion stage, we aggregate low-level detailed information with high-level global information through a jump connection structure and Fusion Block specially designed for this purpose. Besides, during the feature fusion stage, we utilize a stacking configuration consisting of up blocks, Small Block (or Fusion Block), and Wide Block in an interleaved manner to accomplish fine feature recovery and feature fusion. In the final up-sampling step, we develop a simple Prediction Head to map the up-sampled features to segmentation output.
\subsubsection{Feature Extraction Module}
This section presents a detailed description of two sampling blocks (Down Block and Up Block), two feature extraction blocks (Small and Wide Block), Fusion Block and the Prediction Head.

\begin{figure}[h]
\centering
\includegraphics[width=0.8\textwidth]{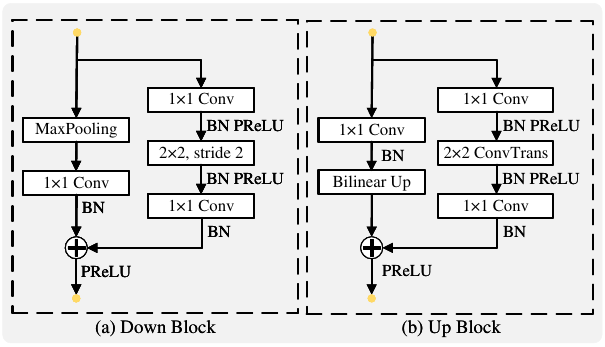}
\caption{The architecture of the Down Block and Up Block.}\label{fig3}
\end{figure}

\textbf{Down Block and Up Block: }To address the issue of vanishing or exploding gradients that may arise in deep neural networks, we incorporate a residual connection architecture within the Down Block. As input and output sizes vary, both branches necessitate sampling during down-sampling. Therefore, we apply point-wise convolution to condense the primary channel, followed by a 2x2 convolution with a stride of 2 for down-sampling the feature map, and then another point-wise convolution to expand the channel count. Meanwhile, the residual channel utilizes max-pooling for down-sampling. To integrate the distinct information from both branches, we merge their sampled outputs and apply the PReLU (Parametric Rectified Linear Unit) activation function, yielding the final output of the sampling module. Similarly, two prevalent methods for up-sampling are interpolation up-sampling and deconvolution. Therefore, we use these two methods to construct the Up Block in the same way and implement channel compression through point-wise convolution before deconvolution. Figure 3 illustrates the specific architecture of these processes. Besides, the PReLU activation function is expressed as follows:

\begin{figure}[h]
\centering
\includegraphics[width=0.6\textwidth]{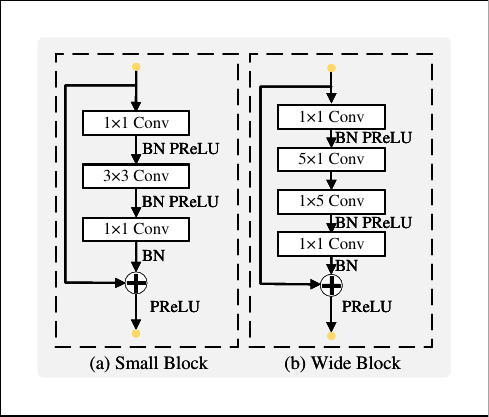}
\caption{The architecture of the Small Block and Wide Block.}\label{fig4}
\end{figure}

\textbf{Small Block and Wide Block:} To adapt to different needs, we devise two unique feature extraction modules with varying receptive field sizes. The first module, dubbed Small Block, consists of a depthwise separable convolution between two point-wise convolutions and employs a residual connection. This module, with a 3x3 receptive field, is optimized for computational efficiency. The second module, termed Wide Block, adopts factorized convolution (5x1 and 1x5) instead of the traditional 5x5 convolution, enabling a larger 5x5 receptive field. Analogous to the Light Block, the Wide Block is flanked by two point-wise convolutions and incorporates a residual connection. Figure 4 showcases the specific architectures of these feature extraction modules. Our proposed feature extraction module exhibits a symmetric channel structure, maintaining an equal number of input and output channels. The initial point-wise convolution reduces the channel count to 1/4 of the output channels, followed by depthwise separable convolution or factorized convolution with an equal number of input and output channels to expand the receptive field. Subsequently, the latter point-wise convolution increases the channel count to achieve the desired output channel dimension. This channel design effectively mitigates the computational complexity arising from large convolution kernels. Considering feature extraction at multiple scales, our module is designed to accommodate receptive fields of 3x3 and 5x5. By utilizing various stacking configurations of feature extraction modules during different stages of down-sampling (Detail Extraction and Semantic Extraction), we can effectively control the receptive field size for each pixel in the feature map at different stages. This approach enables efficient extraction of both low-level detailed information and high-level global information according to our requirements. 

\begin{figure}[h]
\centering
\includegraphics[width=0.6\textwidth]{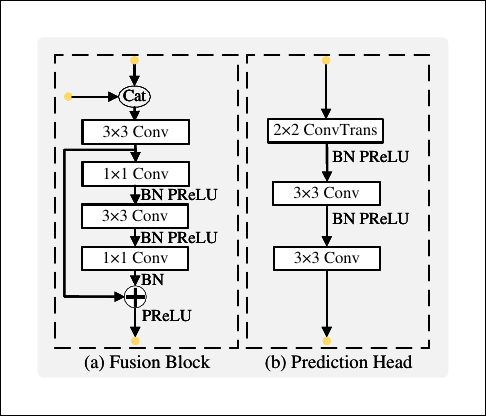}
\caption{The architecture of the Fusion Block and Prediction Head.}\label{fig5}
\end{figure}

\textbf{Fusion Block:} In the up-sampling process, we fuse low-level detailed displacement with high-level global displacement after up-sampling through the Fusion Block at 1/4 size of input, as it is the boundary between the detail extraction stage and the global extraction stage. In the Fusion Block, we concatenate the feature map obtained from up-sampling high-level global information with the details extracted during the detail extraction stage. Subsequently, lightweight depthwise separable convolution are utilized to compress the channel count and integrate the spatial information across different channels. This combined feature map is then fed into the Small Block for further feature fusion and extraction operations. Figure 5(a) illustrates the detailed architecture of this process.

\textbf{Prediction Head:} In the final up-sampling stage of our network, we have designed a straightforward Prediction Head. This structure consists of a deconvolution layer, a point-wise convolution layer, and a 3x3 standard convolution layer, as depicted in Figure 5(b). The deconvolution layer is responsible for up-sampling the feature map, initially half the size of the input image, while simultaneously reducing the number of channels. The point-wise convolution layer serves to integrate spatial information from various channels of the up-sampled feature map. Lastly, the 3x3 standard convolution layer maps the feature map into the desired displacement prediction, thereby completing the entire network computation process.

\begin{table}[h]
\caption{The number of channels in each stage of Bayes-DIC Net.}
\label{table}
\setlength{\tabcolsep}{3pt}
\begin{tabular}{p{\columnwidth}}
$\includegraphics[width=\columnwidth]{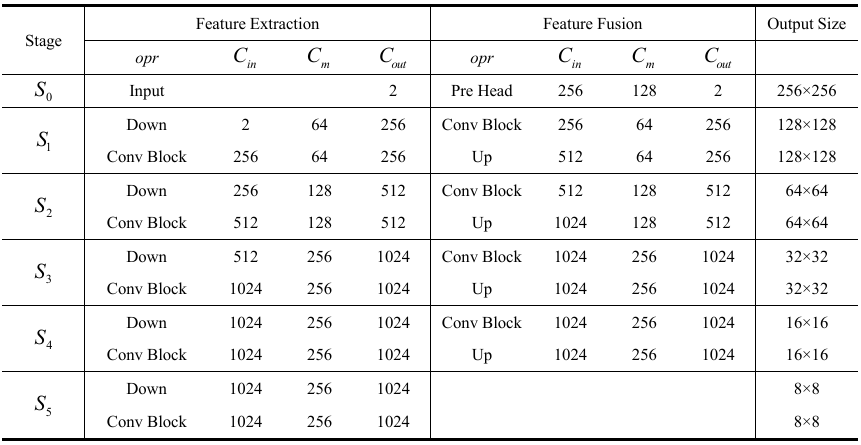}$
\end{tabular}
\label{table1}
\end{table}

Besides, Table 1 provides further details of the module setting, where the \textit{opr} represents different operations at different stages, the Input represents input image, the Down and the Up represent Down Block and Up Block respectively, the Conv Block represents feature extraction module (including (Bayes-)Small Block, (Bayes-)Wide Block and Fusion Block), the ${C_{in}}$ represents the number of input channels, the ${C_m}$ represents the number of intermediate channels, the ${C_{out}}$ represents the number of output channels, and the Output Size represents the resolution of output feature graph of each module.

\subsection{Bayesian DIC Networks}
Bayesian neural networks have the characteristic of providing confidence in their own prediction results, which can make the network's prediction results more convincing. Therefore, it is of great significance for DIC algorithms. To create Bayesian neural networks, Dropout is implemented following convolution layers in numerous approaches[......]. Building upon these studies, we have incorporated the approach of creating Bayesian neural networks into our DIC network construction process, which will be elaborated on in detail in the following passages.

In order to construct a Bayesian convolutional neural network and enable our proposed DIC network to provide predictive confidence, we added a Dropout module into the feature extraction modules of Bayes-DIC-Net. When there are multiple Bayesian modules in a neural network, it can become a Bayesian neural network.

\begin{figure}[h]
\centering
\includegraphics[width=0.6\textwidth]{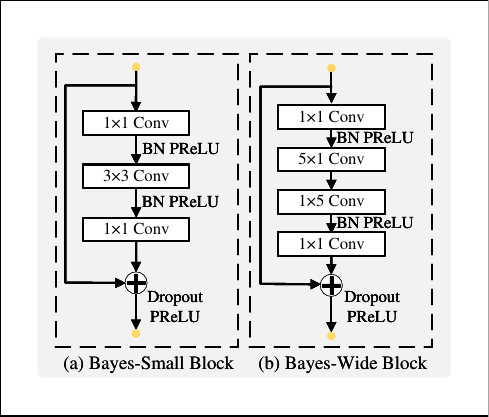}
\caption{The architecture of the Bayes-Small Block and Bayes-Wide Block.}\label{fig6}
\end{figure}

More specifically, we first added a Dropout module to all convolutional operation modules in Bayes-DIC Net and set the dropout probability to 0.1 (As shown in Figures 6). The Dropout module randomly discards some weights during the training process, thereby reducing the risk of model overfitting. In the inference stage, we will manually activate all dropout modules and copy the images used for testing into 8 copies, that is, input each image into a set of 8 identical images. Due to the existence of the dropout module, the trained model will provide different displacement field predictions for each image, that is, eight different displacement field predictions. By averaging the 8 predicted results, we can obtain the final predicted displacement field for that input. Meanwhile, when we calculate the variance of the displacement values at each point of the 8 predicted displacement fields, we can obtain the variance map of the displacement field prediction.

According to the properties of Bayesian neural network, areas with greater variance indicate lower confidence in the network's results. Consequently, we can use the obtained variance map to qualitatively assess the network's confidence level when predicting unlabeled data. This enhances the credibility of the network's predictive outcomes and also assists in real-time evaluation of the network's suitability for detection tasks.

\section{Experiments}

In this section, we first introduce the dataset used in the experimental process, our experimental setup, and the evaluation metrics used. Then, we compared and analyzed the performance and network structure brightness of our proposed method compared to other state-of-the-art algorithms on the generated dataset and the data collected during the experimental process.

\subsection{Datasets, Settings, and Evaluation Metrics}

\subsubsection{Datasets}
Generation Dataset: The whole dataset is generated through the method in this paper. There are 12500 pairs of images with 256×256 resolution in the dataset, and 10000 pairs are training set while the other 2500 pairs are test set.

DIC-Bank Dataset: There are two set of images in this dataset. The Sample 1 is the uniaxial compression test of the reinforced concrete wall, while the Sample 2 is the crack propagation in the rock under compression. There are 40 images in the Sample1 and 10 images in the Sample 2. Besides, for neural networks inference, we crop all the images to 256×256 resolution. During the testing phase, the first and second images are used as a set of inputs to obtain a set of outputs.

\subsubsection{Settings}

\textbf{Training:} To ensure fairness, all models are trained from scratch. We employ the Adam algorithm with a learning rate of 0.0001. For the training stage, we adopt a batch size of 12 and train all networks for 1000 epochs during training stage.

\textbf{Data augmentation:} There is no additional data augmentation.

\textbf{Evaluation:} When testing network performance, we employ the simplest and fastest method, which involves directly loading the test data to assess the performance of every model after training.

\textbf{Setup:} We conduct experiments using PyTorch 2.2.1, and all models are evaluated on a single NVIDIA GeForce GTX 4090 with CUDA 11.8 and CUDNN 8.9.7.

\subsubsection{Evaluation Metrics}
In this paper, we use the Avg. Error to present the average absolute error for the whole dataset and the Max Avg. Error to present the max average absolute error for a pair of images.

\begin{table*}[htbp]
    \centering 
    \caption{Quantitative Comparison with State-of-arts Methods}
    \begin{tabularx}{\textwidth}{
        >{\hsize=1.14\hsize\centering\arraybackslash}X 
        >{\hsize=0.81\hsize\centering\arraybackslash}X 
        >{\hsize=1.14\hsize\centering\arraybackslash}X 
        >{\hsize=0.79\hsize\centering\arraybackslash}X 
        >{\hsize=1.12\hsize\centering\arraybackslash}X 
    }
        \toprule 
        \textbf{Methods} & \textbf{\textit{Avg. Error u}} & \textbf{\textit{Max Avg. Error u}} & \textbf{\textit{Avg. Error v}} & \textbf{\textit{Max Avg. Error v}} \\
        \midrule 
        FlowFormer & 1.3103 &2.5248 & 1.3185 & 2.7136 \\
        RAFT & 1.5308 & 2.8163 & 1.5342 & 3.1003 \\ 
        PWCNet & 0.0236 & 0.3168 & 0.0230 & 0.5855 \\ 
        SeparableFlow & 0.1438 & 0.1777 & 0.8121 & 0.9946 \\
        U-Net & 0.0263 & 0.2178 & 0.0252 & 0.3996 \\
        DIC-Net-d & 0.0208 & 0.1434 & 0.5268 & 0.2983 \\
        Displacement-Net & 0.0140 & \textbf{0.1203} & 0.0138 & 0.2432 \\ 
        \midrule 
        Bayes-DIC Net & \textbf{0.0112} &0.1242 & \textbf{0.0124} & \textbf{0.2393} \\
        \bottomrule 
    \end{tabularx}
    \label{results_table}
\end{table*}

\subsection{Comparative Experiments on Generation Dataset}

\begin{figure}[h]
\centering
\includegraphics[width=1\textwidth]{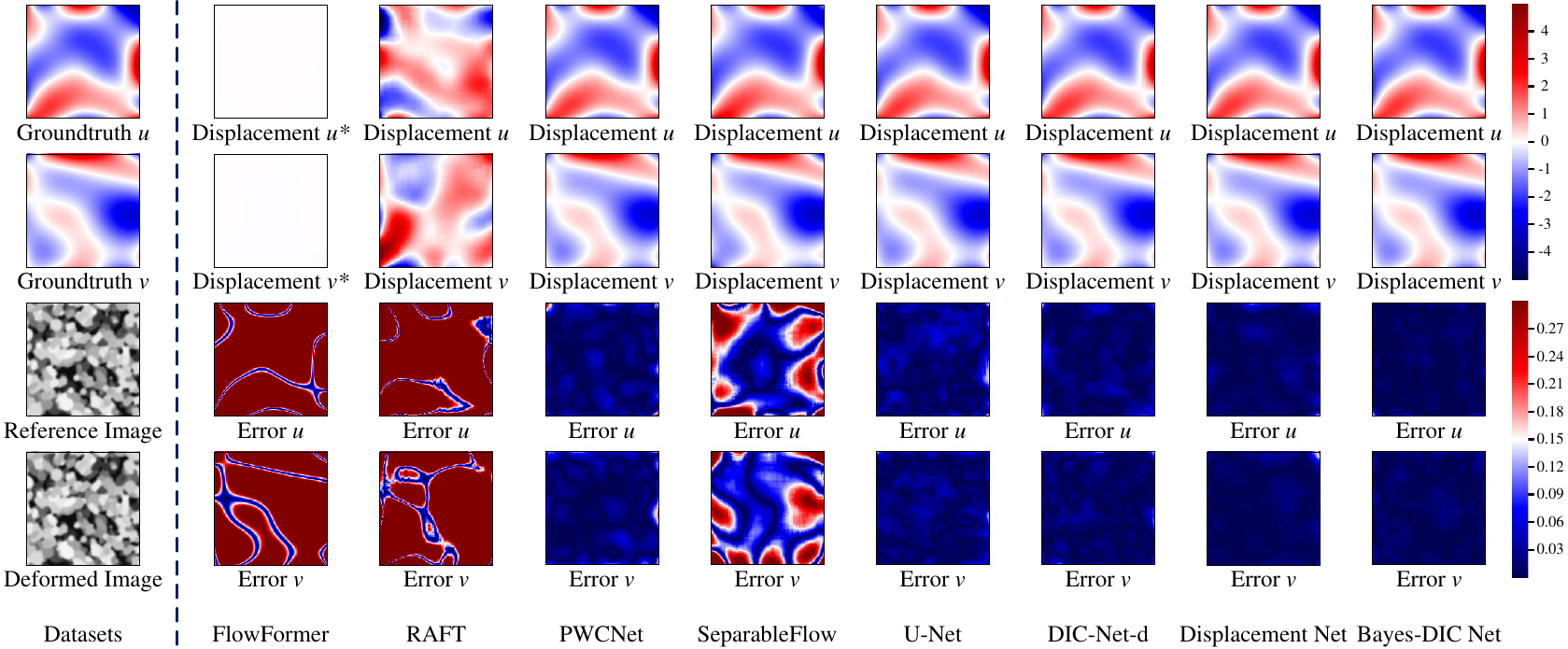}
\caption{The visual display of the prediction and error map of networks on the Generation Dataset.}\label{fig7}
\end{figure}

\begin{figure}[h]
\centering
\includegraphics[width=1\textwidth]{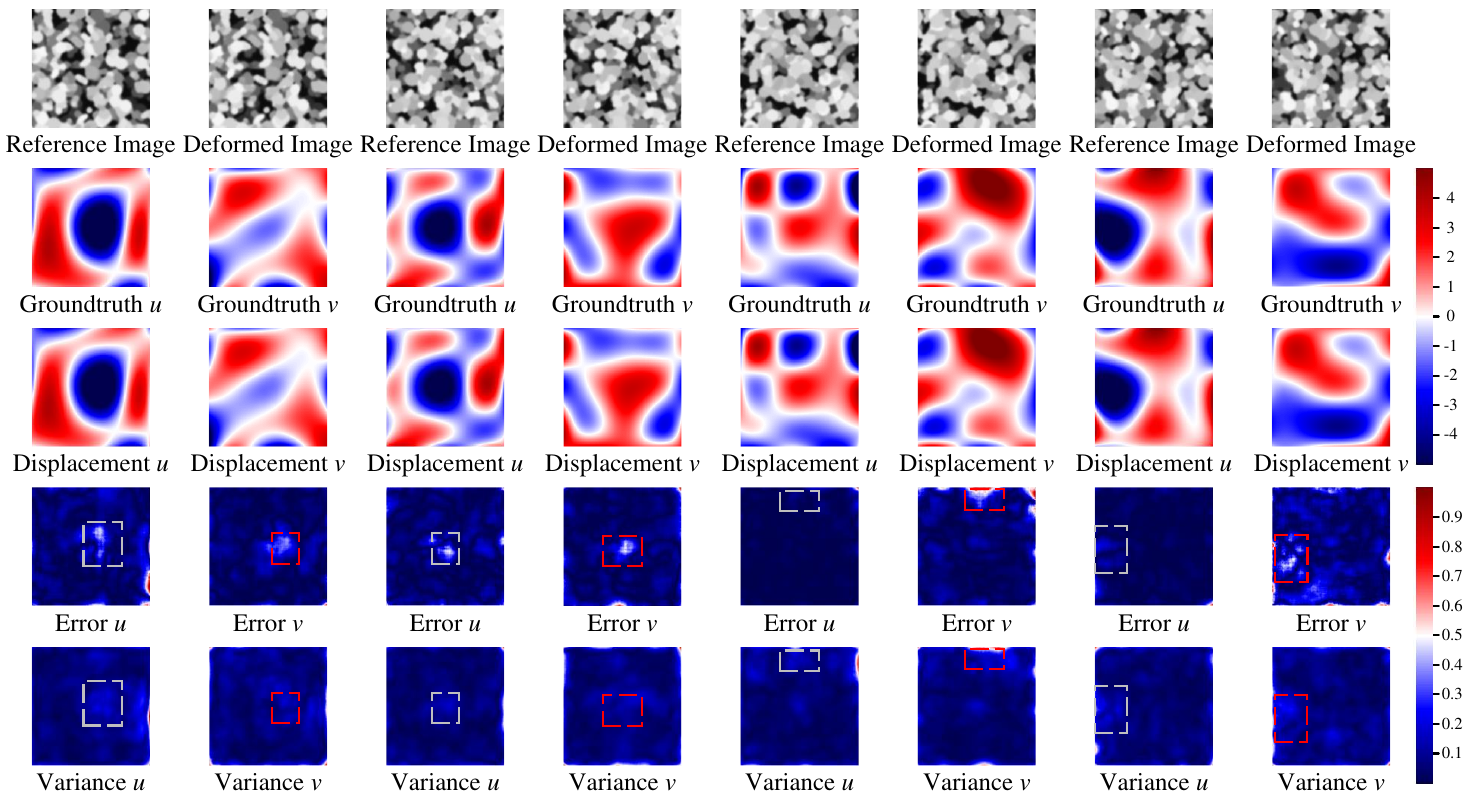}
\caption{The visual display of the prediction, error map and variance map of Bayes-DIC Net on the Generation Dataset.}\label{fig8}
\end{figure}

We ended up choosing eight networks (FlowFormer \cite{huang2022flowformer}, PWCNet \cite{sun2018pwc}, SeparableFlow \cite{zhang2021separable}, Flownet2 \cite{ilg2017flownet}, RAFT \cite{teed2020raft}, U-Net \cite{ronneberger2015u}, DIC-Net-d \cite{wang2023dic} and Displacement-Net \cite{yang2022deep})  as the baseline network to compare with our network.

Table 2 shows the performance of each baseline network and A-Net proposed in this article on the dataset generated in this article. Bold numbers indicate that the network ranks first in this metric.

From Avg. Error's perspective, our proposed Bayes-DIC Net achieved the minimum average error in the u and v directions on the dataset. And compared to the second place data, the network testing structure proposed in this article has decreased Avg values in the u and v directions by 20\% and 10.1\%, respectively. This indicates that our proposed network has good displacement detection and generalization capabilities, and can accurately detect the displacement in different directions at any position in the image after deformation. This proves that our proposed network has good application prospects. 

From the perspective of Max Avg. Error, our proposed Bayes-DIC Net also performs well, only slightly lagging behind Displacement Net in the u direction, but at the same time far outperforms the performance of other baseline networks. In the v direction, our proposed network is slightly higher than the second ranked Displacement Net, but much higher than other compared networks. These results indicate that our proposed network has high prediction stability and will not experience significant prediction bias, which would significantly reduce its reliability.

Figure 7 displays the visual outputs of each comparative network on the dataset, where the asterisk (*) indicates that the predicted output of this network is close to zero, so the visualization of the predicted output appears as white. From the displacement map of the predicted output, it can be seen that the predicted results of our proposed network are almost identical to the true values, indicating that our proposed network has extremely high predictive performance. From the error map of the predicted output, compared with the error maps of other networks, the error map of the network proposed in this paper basically does not have red areas (areas with high error), and the color of the error map is darker, indicating that its predicted output maintains a small error compared to the true value in the entire image, thus once again proving the extremely high performance of the network proposed in this paper.

Figure 8 displays the visual normalized error map and normalized variance map of the Bayes-DIC Net's outputs. Besides, due to the large outliers in the values of the variance map in the boundary and vertex regions, we truncated the variance map before the normalization operation. All variance data greater than 0.02 were set to 0.02 to eliminate the influence of large values in small regions on the visualization, making the visualization results more beautiful and reflecting richer information. It can be clearly seen from the visualization results that the areas with large output variance of the model can correspond to the areas with relatively large errors, indicating that they have a certain connection. This indicates that when the model obtains the predicted value and its variance, we can use the variance to determine whether our predicted value is accurate. In this case, when there is no true value, we can roughly judge the confidence of the network output by the variance of the network output. Therefore, the Bayes DIC Net proposed in this paper will have greater advantages and potential in practical applications.


\section{Conclusion}
The primary contributions of this paper can be summarized into the following three points:

High-Quality DIC Dataset Generation Method: We introduce a DIC dataset generation method based on non-uniform B-spline surfaces. By randomly generating control point coordinates, we are able to create displacement fields that encompass various real displacement scenarios, which are then used to generate speckle pattern datasets. This approach enables us to produce a large-scale dataset that captures as many real displacement field situations as possible, thereby enhancing the training of deep learning-based DIC algorithms and improving their generalization capabilities.

Innovative Network Architecture Bayes-DIC Net: We propose a novel network architecture named Bayes-DIC Net, which extracts information at different levels in stages during the down-sampling phase and facilitates the aggregation of information across various levels through a single skip connection in the up-sampling phase. Additionally, a series of lightweight convolution blocks are designed for Bayes-DIC Net, which expand the receptive field and capture rich contextual information while minimizing computational costs.

Application of Bayesian Neural Networks: By incorporating appropriate dropout modules into the desianed network and activating them during the network inference stage, Bayes-DIC Net transforms into a Bayesian neural network. This network, when processing real unlabeled datasets, is capable of providing not only prediction results but also confidence levels in these predictions. This feature significantly enhances the practicality and reliability of our network in real displacement field prediction tasks and reduces user distrust in the algorithm.

As deep learning and computer vision technologies continue to evolve, the potential applications of Digital Image Correlation (DIC) in fields like materials science, biomechanics, and structural engineering are immense. Although the methods outlined in this paper have established a robust foundation, there remain significant challenges and opportunities for future research. For instance, enhancing the diversity and scale of datasets is essential; future efforts could expand datasets to include a wider range of materials, lighting conditions, and complex backgrounds, thereby improving algorithm generalization.

\section*{Acknowledgements}
Thanks to Professor Ping Guo from Department of Mechanical Engineering, Northwestern University for his valuable guidance, insightful suggestions, and continuous support throughout the course of this research.

\appendix



\printbibliography






\end{document}